\documentclass{article}
\pdfoutput=1 %

\usepackage[preprint]{neurips_data_2024}

\usepackage[utf8]{inputenc} %
\usepackage[T1]{fontenc}    %
\usepackage{hyperref}       %
\usepackage{url}            %
\usepackage{booktabs}       %
\usepackage{amsfonts}       %
\usepackage{nicefrac}       %
\usepackage{microtype}      %
\usepackage{xcolor}         %
\usepackage{multirow}
\usepackage{makecell}
\usepackage{comment}
\usepackage{wrapfig}
\usepackage{amsmath}
\usepackage{enumitem}
\usepackage{arydshln}
\usepackage[most]{tcolorbox}
\usepackage{subcaption}
\usepackage{graphicx}
\usepackage{authblk}
\usepackage{hyperref} %

\setlist[itemize]{leftmargin=2em}
\setlist[enumerate]{leftmargin=2em}

\newtcolorbox{prompt}[1]{
    enhanced,
    colback=gray!20,
    colframe=black,
    boxrule=0.5pt,
    arc=3mm,
    left=10pt,
    right=10pt,
    boxsep=5pt,
    fonttitle=\bfseries,
    title=#1,
}

\title{MASSW: A New Dataset and Benchmark Tasks for AI-Assisted Scientific Workflows}

\makeatletter
\renewcommand\AB@affilsepx{\quad\protect\Affilfont}
\makeatother
\renewcommand*{\Affilfont}{\normalsize\normalfont}

\newcommand{\customdashline}{
  \hdashline
  \noalign{\vskip 0.5ex}
}
\author[1*]{Xingjian Zhang}
\author[1*]{Yutong Xie}
\author[1]{Jin Huang}
\author[2]{Jinge Ma}
\author[2]{Zhaoying Pan}
\author[1]{Qijia Liu}
\author[1]{Ziyang Xiong}

\author[3]{Tolga Ergen}
\author[3]{Dongsub Shim}

\author[1]{Honglak Lee}
\author[1]{Qiaozhu Mei}

\affil[1]{University of Michigan, Ann Arbor}
\affil[2]{Purdue University}
\makeatletter
\renewcommand\AB@affilsepx{\\\protect\Affilfont}
\makeatother
\affil[3]{LG AI Research}

\affil[*]{\textit{Equal Contribution}}
\affil[1]{\texttt {\{jimmyzxj,yutxie,huangjin,ponypony,xziyang,qmei\}@umich.edu\quad honglak@eecs.umich.edu}}
\affil[2]{\texttt{\{ma859, pan433\}@purdue.edu}}
\affil[3]{\texttt{\{tergen, dongsub.shim\}@lgresearch.ai}}

\begin{document}

\maketitle

\begin{abstract}
Scientific innovation relies on detailed workflows, which include critical steps such as analyzing literature, generating ideas, validating these ideas, interpreting results, and inspiring follow-up research. However, scientific publications that document these workflows are extensive and unstructured. This makes it difficult for both human researchers and AI systems to effectively navigate and explore the space of scientific innovation. To address this issue, we introduce \textbf{MASSW}, a comprehensive text dataset on \textbf{M}ulti-\textbf{A}spect \textbf{S}ummarization of \textbf{S}cientific \textbf{W}orkflows. MASSW includes more than 152,000 peer-reviewed publications from 17 leading computer science conferences spanning the past 50 years. Using Large Language Models (LLMs), we automatically extract five core aspects from these publications -- \emph{context}, \emph{key idea}, \emph{method}, \emph{outcome}, and \emph{projected impact} -- which correspond to five key steps in the research workflow. These structured summaries facilitate a variety of downstream tasks and analyses. The quality of the LLM-extracted summaries is validated by comparing them with human annotations. We demonstrate the utility of MASSW through
multiple novel machine-learning tasks that can be benchmarked using this new dataset, which make various types of predictions and recommendations along the scientific workflow. MASSW holds significant potential for researchers to create and benchmark new AI methods for optimizing scientific workflows and fostering scientific innovation in the field. Our dataset is openly available at \url{https://github.com/xingjian-zhang/massw}.
\end{abstract}

\section{Introduction}
\label{sec:introduction}

Can AI be a capable copilot for scientific research? Scientific innovation is driven by complex and detailed  workflows, also referred to as scientific methods at a coarse level \citep{ayala2009darwin, voit2019perspective}. These workflows typically involve critical steps such as analyzing existing literature, generating novel ideas, validating these ideas through analyses and experiments, interpreting experimental results, and ultimately inspiring future research inquiries. To navigate and explore the space of innovations, both the pilot and the copilot have to understand, plan, and optimize the scientific workflows \citep{wang2023scientific}.  These workflows are often documented in scientific publications, which are crucial for scientists to understand and reproduce existing research, as well as plan and accelerate new research. However, the traditional format of these publications is unstructured and complex, which does not readily facilitate efficiently tracing scientific workflows and extending them towards new scientific research. To assist researchers in better navigating and exploring the scientific innovation space, it is essential to develop new datasets that document scientific workflows in a more structured way, along with new tools to reason through and evolve these workflows.

Developing scientific workflow datasets is challenging. While human experts are skilled at deciphering complex scientific publications, their highly personalized interpretations, if not sufficiently aligned, often lead to inconsistent and heterogeneous annotations and predictions~\citep{Beck2020-dh}.
Furthermore, annotations by highly specialized researchers are inherently expensive, limiting their applicability in building large datasets at the scale and scope of a scientific field~\citep{Takeshita2024-sv,Fisas2015-ai,Cachola2020-dg, mei2008generating}.
These challenges highlight the need for an automated, scalable, and consistent solution to annotate structured scientific workflows, a task well-suited for an AI.  Indeed, the advent of large language models (LLMs) has demonstrated promising performance in reasoning through natural language \citep{10.5555/3600270.3602070}, positioning them as a viable candidate for automating the annotation of scientific workflows, even though it remains to be seen whether they can match the accuracy of human experts.

Addressing these challenges, we present \textbf{MASSW}, a novel and large-scale dataset that provides a comprehensive and structured \textbf{M}ulti-\textbf{A}spect \textbf{S}ummarization of \textbf{S}cientific \textbf{W}orkflows. The key features of MASSW include
\begin{itemize}
    \item \textbf{Structured scientific workflows}. MASSW defines five core aspects of a scientific workflow -- \emph{context}, \emph{key idea}, \emph{method}, \emph{outcome}, and \emph{projected impact}. These aspects align with the typical stages in general scientific workflows that can be identified in the literature. Utilizing LLMs, we are able to extract and structure these five aspects from each publication with consistency.
    \item \textbf{Large scale}. MASSW contains the structured scientific workflows and meta-information from over 152,000 peer-reviewed publications, across 17 leading computer science conferences, and spanning the past 50 years. 
    \item \textbf{Accuracy}. The coverage and accuracy of MASSW have been validated through comprehensive inspections and comparisons with human annotations and alternative methods.
    \item \textbf{Rich benchmark tasks}. MASSW facilitates multiple novel and benchmarkable machine-learning tasks, such as idea generation and outcome prediction. It supports diverse tasks centered on predicting, recommending, and expanding key elements of a scientific workflow, serving as a benchmark for evaluating LLM agents' ability to navigate scientific research.
\end{itemize}
By providing a large-scale, structured, and accurate resource, MASSW opens new avenues for researchers to develop and evaluate innovative AI methods that facilitate more effective scientific workflows, fostering greater and faster innovations within the field.

\section{Dataset Overview}  %
\label{sec:dataset_overview}
MASSW is a \textit{structured} and \textit{large-scale} dataset designed to enhance the exploration and analysis of scientific workflows.  
In Section \ref{sec:core-aspects}, we first discuss how to structure a scientific publication into five core aspects, corresponding to five key steps in a general scientific research workflow. 
In Section \ref{sec:data-curation}, we describe the curation of scientific publication data and an automated procedure that summarizes these core aspects with LLMs.  
Lastly, we present basic statistics about the constructed MASSW dataset and a multi-view visualization of these aspects in Section \ref{sec:statistics-and-vis}. 

\subsection{Core Aspects of Scientific Workflows}
\label{sec:core-aspects}

\begin{table}[t]
\centering
\def\arraystretch{2.0}%
\begin{tabular}{p{2.1cm}p{5.5cm}p{5cm}}
\toprule
\textbf{Aspect} & \textbf{Definition} & \textbf{Example}  \\ \midrule
\textbf{Context} \newline {\small \textit{Ask questions, review literature \newline (prior to study)}} & The status quo of related literature or reality which motivated this study. This could normally be a problem, a research question, or a research gap that has not been successfully addressed by previous work. & \textit{Making language models bigger does not inherently make them better at following a user's intent, as large models can generate outputs that are untruthful, toxic, or not helpful.} \\
\textbf{Key Idea} \newline {\small \textit{ Construct hypothesis (proposed in this study)}} & The main intellectual merit of this paper, often in comparison to the context. This could normally be a novel idea or solution proposed in this paper that distincts it from what’s already done in literature. & \textit{The authors propose InstructGPT, a method to align language models with user intent by fine-tuning GPT-3 using a combination of supervised learning with labeler demonstrations and reinforcement learning from human feedback.} \\
\textbf{Method} \newline {\small \textit{Test hypothesis \newline (after hypothesis construction)}} & The specific research method that investigates and validates the key idea. This could be an experimental setup, a theoretical framework, or other necessary validation methodology to implement and/or evaluate the key idea. & \textit{The authors evaluate the performance of InstructGPT by humans on a given prompt distribution and compare it with a much larger model GPT-3.} \\
\textbf{Outcome} \newline {\small \textit{Interpret results, draw conclusion \newline (after testing hypothesis)}} & The factual statement about the study output. This could be the experiment results and any other measurable outcome that has occurred. It marks whether the key hypothesis is testified or not. & \textit{InstructGPT, even with 100x fewer parameters, is preferred over GPT-3 in human evaluations. It shows improvements in truthfulness and reductions in toxic outputs with minimal performance regressions on public NLP datasets.} \\
\textbf{Projected \newline Impact} \newline {\small \textit{Future work \newline (anticipated but not yet done)}} & The author-anticipated impact of the work on the field, and potential further research identified by the author that may improve or extend this study. & \textit{Fine-tuning with human feedback is a promising direction for aligning language models with human intent.} \\ \bottomrule
\end{tabular}
\vspace{1em}
\caption{Core aspects in the MASSW dataset that correspond to key steps (\textit{in italic}) in a general scientific workflow. The example is based on the paper ``Training Language Models to Follow Instructions with Human Feedback.''~\citep{Ouyang2022-hl} More examples of MASSW are provided in Appendix~\ref{sec:example-aspects}.}
\label{tab:study_framework}
\end{table}

A typical workflow of scientific research often involves common steps: asking a general research question and reviewing existing literature, formulating a hypothesis or research idea, validating the hypothesis, interpreting the results and drawing conclusions, reporting the findings, and planning follow-up research~\citep{ayala2009darwin, voit2019perspective}. A scientific publication often describes some or all of these steps with corresponding narrative aspects. For example, the authors often motivate their study within the \emph{context} of existing research, highlight the \emph{key idea} of the study, describe the \emph{method} used to validate their idea, discuss the \emph{outcome} of the validation, and articulate the \emph{potential impact} of the study from the author's perspective.
In Table \ref{tab:study_framework}, we define these core aspects more formally. 

\paragraph{Context} The context of a study summarizes the status quo of the research field or the broader reality before the study is presented.
This aspect is often related to analyzing relevant literature, identifying the gap and unresolved challenges, and motivating new research ideas to fill the gap. 
In a scientific publication, this key aspect is often described as \emph{background}, \emph{challenges}, or \emph{literature review}, as approached by previous work of publication summarization \citep{Fisas2015-ai,Takeshita2024-sv}. 

\paragraph{Key Idea} The key idea represents the central hypothesis or novel contribution proposed in the study. This is the key aspect that distinguishes the current work from the context of existing work. It is a product of idea generation, a critical step in the scientific workflow where new concepts are formed, new connections are made, and new solutions are proposed to address particular challenges in research. In previous work of text summarization, it is sometimes related to the \emph{approach} described in a paper, which only partially reflects its key ideas  \citep{Fisas2015-ai,Takeshita2024-sv}. 

\paragraph{Method} The method of a study details the procedures and techniques used to validate the key idea or hypothesis. In other words, the method is not a part of the hypothesis itself, but rather the procedure used to prove or reject the hypothesis. In previous work of text summarization, \textit{method} is sometimes confused with the \textit{key idea} (both referred to as part of the \emph{approach} \citep{Fisas2015-ai,Takeshita2024-sv}), especially when the main subject of the research is a ``method.'' We explicitly distinguish \textit{method} from the \textit{key idea} as they refer to different steps in the scientific workflow (generating ideas v.s. validating ideas). 

\paragraph{Outcome} The outcome includes the results and findings as a product of the \textit{method} in the study.
This aspect corresponds to the measurable results, the interpretation of these results, and other type of impact of the work that has already happened by the time of publication. This concept is also mentioned in previous work of text summarization, as ``outcome'' or ``result'' \citep{Fisas2015-ai,Takeshita2024-sv}. 

\paragraph{Projected Impact} The projected impact outlines the potential future implications of the research that has not happened at the time of publication. This aspect is often an ex-ante prediction of how the results of the work would inspire follow-up research or deployment, from the author's point of view. It discusses how the findings can contribute to the field, suggest new research directions, and potentially lead to societal or technological advancements. Previous work often simply uses the concept of \emph{future work}, while ignoring the broader impact of the study \citep{Fisas2015-ai,Takeshita2024-sv}. 

\subsection{Data Curation and Aspect Summarization}
\label{sec:data-curation}

Advancing AI's understanding  and ability to improve scientific workflows requires large-scale and high-quality data. To address this challenge, we curate a collection of scientific publications and structure it into the above-defined core aspects at scale. 

\paragraph{Large-scale scientific publication collection. }

To build this initial version of the MASSW dataset, we focus on Computer Science publications from 17 top-tier conferences listed in \href{CSRankings.org}{CSRankings.org}, which we identify as relevant to the broader field of AI. 
We access the publications through Open Academic Graph (OAG)\footnote{The OAG dataset is publicly released under the ODC-BY license.}, a linked graph database for academic entities including publications, venues, affiliations, and authors
\citep{zhang2022oag,zhang2019oag}. 
In total, 191,055 papers are collected that span from 1969 to 2024, among which, 152,027 contain both a title and an abstract. 
More details about data curation can be found in Appendix \ref{app:data-curation}.

\begin{wraptable}{r}{0.6\textwidth}
    \centering
    \vspace{-10pt}
    \begin{tabular}{rrr}
        \toprule 
         & \#Papers with & Avg. \#Tokens \\
        \midrule 
        Abstract & 152,027 & 145.3 \\
        \customdashline
        Context & 149,849 & 34.8 \\
        Key Idea & 149,411 & 35.1 \\
        Method & 142,241 & 30.7 \\
        Outcome & 132,614 & 27.6 \\
        Projected Impact & 72,983 & 27.2 \\
        \customdashline
        All Aspects & 62,506 & N/A \\
        \bottomrule
    \end{tabular}
    \caption{Basic statistics of MASSW. }
    \vspace{-10pt}
    \label{tab:statistics}
\end{wraptable}

\paragraph{Automatic aspect summarization with LLMs. }

Most relevant datasets on structured summary of publications were created using human annotations, which only cover tens to thousands of papers \citep{mei2008generating, Fisas2015-ai,cachola-etal-2020-tldr,wang-etal-2022-squality,Takeshita2024-sv}.  %
For MASSW, we leverage the power of LLMs (e.g., GPT-4) to automatically summarize the five core aspects of the entire set of collected papers that have a title and an abstract. 
More details of LLM-based summarization, including the prompts used, are described in Appendix \ref{app:aspect-summ}. 

\subsection{Dataset Statistics and Visualization}
\label{sec:statistics-and-vis}

Table \ref{tab:statistics} reports basic statistics of the MASSW dataset and per-aspect summaries. We also include a multiview visualization in Figure \ref{fig:largevis-aspects} in Appendix \ref{app:data-vis} to give an overview of the core aspects.

\section{Dataset Validation}
\label{sec:dataset_validation}
Are LLM-generated summaries trustfully describing the core aspects of the scientific workflow? We validate the structured summaries in the MASSW dataset by comparing them with human-generated summaries. We have curated a small-scale subset of publications and solicited the summaries of the same five aspects from human annotators. This subset demonstrates the alignment between the LLMs and human experts in generating the multi-aspect summary of scientific workflows.

\subsection{Evaluation Metric}

We employ two categories of similarity evaluation metrics: lexical-level and semantic-based. Lexical-level metrics, such as \textit{BLEU}~\citep{Papineni2002-ab} and \textit{ROUGE}\footnote{We report ROUGE-1 that evaluates on unigram.}~\citep{Lin2004-yl}, are prevalent across various natural language generation tasks. However, numerous studies \citep{Sellam2020-en, Callison-Burch2006-xd, Ananthakrishnan2006-sd, Sai2022-hy} indicate their limited alignment with human judgments, primarily due to their reliance on exact word matches. Conversely, semantic-based metrics represent a more advanced approach, assessing the similarity in meaning or content through the use of pre-trained language models. For semantic evaluation, we utilize three metrics: \textit{BERTScore (BS)}~\citep{Zhang2019-ir}, which compares token-wise contextual embeddings, \textit{cosine similarity (CS)}, derived from embeddings generated by~\citep{Wang2024-nf}, and \textit{BLEURT}~\citep{Sellam2020-en}, which is fine-tuned to reflect human judgment. Their implementation details can be found in Appendix~\ref{sec:semantic-metrics}.

\subsection{Human Annotation}

We use a proportionate stratified sampling method on different venues and publication times to select the annotation subset. Specifically, we sample 7 papers from each of the venues in different times, resulting in a total of 126 papers for annotation. This strategy ensures a representative sample across the broad spectrum of AI research. The complete annotation process is detailed in Appendix~\ref{sec:annotation}.

Two trained human experts who are familiar with reading scientific literature are assigned to annotate the aspects of each paper, based on the title and abstract, following a carefully designed codebook.  Table~\ref{table:eval} (top) illustrates the agreement between human experts by treating one annotation as the reference and the other as the prediction for each paper. \footnote{\label{footnote1}To help understand the scale of these metrics, we include a range of examples with varying levels of similarity in Appendix~\ref{sec:example-similarity}} In general, there is a high level of agreement across all aspects of scientific workflow, suggesting that the proposed five aspects are well-defined and the annotations do not have obvious individual bias.

\begin{table}[ht]
\centering
\begin{tabular}{lcccccc} 
\toprule
                                     & Aspects          & CS & BLEURT & BS & BLEU  & ROUGE-1  \\
\midrule
\multirow{5}{*}{\shortstack[l]{Human\\Agreement}}     & Context          & 0.935 & 0.656 & 0.942 & 0.594 & 0.703    \\
                                         & Key Idea         & 0.944 & 0.618 & 0.938 & 0.464 & 0.637    \\
                                         & Method           & 0.900 & 0.559 & 0.924 & 0.357 & 0.540    \\
                                         & Outcome          & 0.936 & 0.671 & 0.950 & 0.608 & 0.737    \\
                                         & Projected Impact & 0.941 & 0.742 & 0.955 & 0.642 & 0.748    \\
\customdashline
\multirow{5}{*}{\shortstack[l]{GPT-3.5-Human\\Alignment}} & Context          & 0.934 & 0.597 & 0.934 & 0.524 & 0.635    \\
                                         & Key Idea         & 0.936 & 0.575 & 0.927 & 0.439 & 0.582    \\
                                         & Method           & 0.895 & 0.510 & 0.910 & 0.197 & 0.445    \\
                                         & Outcome          & 0.928 & 0.608 & 0.934 & 0.452 & 0.626    \\
                                         & Projected Impact & 0.876 & 0.498 & 0.905 & 0.170 & 0.371    \\
\customdashline
\multirow{5}{*}{\shortstack[l]{GPT-4-Human\\Alignment}} & Context          & 0.940 & 0.607 & 0.934 & 0.384 & 0.604    \\
                                         & Key Idea         & 0.944 & 0.582 & 0.928 & 0.375 & 0.572    \\
                                         & Method           & 0.894 & 0.510 & 0.908 & 0.197 & 0.450    \\
                                         & Outcome          & 0.931 & 0.603 & 0.933 & 0.355 & 0.596    \\
                                         & Projected Impact & 0.916 & 0.611 & 0.933 & 0.282 & 0.563    \\
\customdashline
\multirow{5}{*}{\shortstack[l]{Mixtral-8x7B-Human\\Alignment}} & Context          & 0.944 & 0.645 & 0.946 & 0.590 & 0.693    \\
                                         & Key Idea         & 0.949 & 0.636 & 0.943 & 0.556 & 0.662    \\
                                         & Method           & 0.905 & 0.554 & 0.920 & 0.295 & 0.509    \\
                                         & Outcome          & 0.933 & 0.674 & 0.948 & 0.665 & 0.707    \\
                                         & Projected Impact & 0.917 & 0.635 & 0.936 & 0.384 & 0.599    \\
\bottomrule
\end{tabular}
\vspace{0.5em}
\caption{Human agreement (top) and LLM-human alignments (bottom) for the five extracted aspects of scientific workflow. CS stands for cosine similarity and BS stands for BERTScore. Human agreement is calculated with one annotation randomly selected as the reference and the other (2 annotations per paper) as the prediction. For each aspect, the results are computed on the papers that both the human and the LLM consider the aspect as being ``mentioned'' in the abstract.
Please see Appendix~\ref{sec:example-similarity} for a range of examples with varying levels of similarity.}
\label{table:eval}
\end{table}
\begin{table}[ht]
    \centering
    \begin{tabular}{lrrrr}
    \toprule
    Model & Context & Method & Outcome & Projected Impact \\
    \midrule
    GPT-3.5 & 0.000 & 0.105 & 0.364 & 0.346 \\
    GPT-4 & \textbf{0.583} & \textbf{0.421} & \textbf{0.636} & \textbf{0.923} \\
    Mixtral-8x7B & 0.042 & \textbf{0.421} & 0.364 & 0.750 \\
    \bottomrule
    \end{tabular}
    \vspace{0.5em}
    \caption{Recall of ``not mentioned'' aspects as identified by human experts. Higher values indicate lower rates of hallucination. Key idea is not included as it presents in all papers in the annotation set.}
    \label{tab:na}
\end{table}

\subsection{Evaluation of MASSW against Human Experts}

Three LLMs are investigated to build the MASSW dataset: GPT-3.5~\citep{openai2022chatgpt}, GPT-4~\citep{openai2023gpt4}, and Mixtral 8x7B~\citep{jiang2024mixtral}. They are instructed using the same information in the codebook for human annotators, and their generated summaries are evaluated against human annotations, shown in Table~\ref{table:eval}. Ideally, if the LLM perfectly aligns with human experts, the similarity between an LLM-annotation and a human-annotation should be comparable to that between the annotations of two humans.

Indeed, for semantic-based metrics, we only see a small difference between LLM-human alignment and human-human agreement, and this pattern is consistent for all three models. This indicates that the semantics of the core aspects of the scientific workflows captured by the LLM closely mirror those by human experts. For lexical-level metrics, there is a more notable disparity, especially between GPT-4 and human experts. Our inspection suggests that this discrepancy primarily arises because GPT-4 tends to generate abstractive summaries, often rephrasing/refining the content contained in the original paper
whereas human annotators are inclined to directly quote the narratives in original paper whenever appropriate. This extractive approach is inherently more compatible with lexical-level metrics, which favor direct word overlaps.

Not all papers have described all the five aspects, and not all descriptions appear in the abstract. If an aspect is not present in the abstract, an annotator should consider it as \textit{not mentioned}. When compared upon cases where both the human and the LLM consider an aspect as ``mentioned'', all three models perform well, with Mixtral-8x7B being particularly more similar to humans (Table~\ref{table:eval}).
However, upon the scenarios where a human annotator does not identify an aspect from the paper's abstract, the LLMs sometimes do, likely due to the problem of hallucination. From Table~\ref{tab:na}, we note that GPT-3.5 has the highest rate of hallucination, especially on the aspect of \textit{projected impact} that is only mentioned in half of the abstracts. Mixtral-8x7B has a lower rate of hallucination, and GPT-4 has the lowest, correctly identifying 92\% of the missing \textit{projected impact} as ``not mentioned''

Given the desirable alignment with humans and the lowest level of hallucination, we select GPT-4 as the annotator for MASSW, which is also easily accessible to the readers to reproduce our results. %

In conclusion, the MASSW dataset, powered by GPT-4, exhibits a high degree of accuracy and a low hallucination rate in identifying the key aspects of scientific workflows from scientific papers.

\section{Benchmark Tasks}
\label{sec:benchmark_tasks}
There are many opportunities of using AI to optimize the workflow of scientific research. Given the sequential nature of a workflow, there is a lot of potential to design machine learning methods to predict and recommend the next steps in a workflow given the earlier sequence. For example, can AI make predictions about novel ideas given the status quo of literature? Can it recommend an appropriate method to validate an idea, and therefore predict the results of the validation? Can it forecast the future impact of a study and suggest follow-up research? These tasks are compelling, but they have not been benchmarkable due to the lack of large-scale data with accurate labels. The presence of MASSW unleashes this potential. 

In this section, we demonstrate how the MASSW dataset can serve as a foundational resource for various downstream tasks.
We benchmark a few methods for a handful of tasks, mostly off-the-shelf LLMs, and invite the community to explore the greater variety of possibilities. We are particularly curious about the ability of LLMs, as~\citep{Franceschelli2023-cg} contends that LLMs exhibit only a "weak version of novelty" and suggests that their inherent autoregressive nature may inhibit their ability to achieve transformational creativity.
We present these representative tasks in Section~\ref{sec:benchmark_task_def} and detail the experimental setups and the performance of baseline methods in Section~\ref{sec:benchmark_baseline}

\subsection{Task Definitions}\label{sec:benchmark_task_def}

We demonstrate two type of meaningful tasks that uses AI to facilitate scientific workflow, as they are naturally induced from the MASSW dataset:

\begin{itemize}
\item \textbf{Workflow Prediction}: A scientific workflow has a sequence of steps, for example, ``digesting the literature'' $\rightarrow$ ``generating research idea'' $\rightarrow$ ``validating the idea'' $\rightarrow$ ``interpreting the results'' $\rightarrow$ ``planning follow-up research''. An effective AI system should proficiently navigate scientific workflows by extrapolating subsequent steps from preceding ones. Hence, for each key aspect in MASSW, we can task a model to make predictions based on the aspects prior in the sequence:

\begin{itemize}
\item \textsc{Idea Generation:} given the \textit{context} of literature, predict the \textit{key idea} of a new study.
\item \textsc{Method Recommendation:} given the \textit{context} and a \textit{key idea}, suggest a \textit{method} to validate the idea.
\item \textsc{Outcome Prediction:} given the \textit{context}, a \textit{key idea}, and a \textit{method} of validation, forecast the \textit{outcome} of the validation/analysis.
\item \textsc{Future Work Recommendation:} given all other aspects of a study, estimate its \textit{projected impact} and recommend tasks for follow-up studies.
\end{itemize}
\item \textbf{Title Prediction}:
A subsequent step of the research workflow is to publish the results. A powerful AI copilot should be able to enhance writing by recommending appropriate and appealing titles that encapsulate the key elements of a paper. We therefore introduce the task of title prediction, which challenges an AI model to generate a title given all five aspects of a study. 
\end{itemize}

\subsection{Demonstration with Baselines}\label{sec:benchmark_baseline}
We now detail the experimental settings and the performance of our baseline models. 
\begin{itemize}
    \item \textbf{Sampling Methods}: To create the test set, we employ proportionate stratified sampling based on dates of publication; we select 60 papers (with all aspects mentioned) from each venue to ensure broad representation.
    \item \textbf{Baseline Models}: We test GPT-3.5, GPT-4, and Mixtral 8x7B as baseline models.
    \item \textbf{Prompting Methods}: We test zero-shot, zero-shot chain of thought (adding the instruction ``\textit{Let's think step by step}'' to the end of the zero-shot prompt)~\citep{kojima2022large}, few-shot, and few-shot chain of thought~\citep{Wei2022-yt} promptings. The models were provided with (i) definitions of all five aspects as defined in Table~\ref{tab:study_framework}, (ii) all necessary aspects for each task, and (iii) a specific task instruction. Detailed prompting templates and settings are specified in the Appendix~\ref{sec:experiment-details}.
    \item \textbf{Evaluation Metrics}: We evaluate the model outputs using the same metrics described in Section~\ref{sec:dataset_validation}. Due to space limit, \textit{BLEURT} and \textit{ROUGE} are reported in Table~\ref{tab:experiment_combined_results} as examples of semantic-based and lexical-level metrics, while other metrics are reported in Appendix~\ref{sec:supplementary-tables}.
\end{itemize}

\begin{table}[h!]
	\centering
	\scriptsize
	\begin{tabular}{llcccccccccc}
		\toprule
		\multicolumn{2}{c|}{} & \multicolumn{5}{c|}{BLEURT Score} & \multicolumn{5}{c}{ROUGE-1 Score} \\ 
		Model                         & Prompt & Idea           & Method         & Outcome        & Future         & Title          & Idea           & Method         & Outcome        & Future         & Title          \\ \midrule
		\multirow{3}{*}{GPT-3.5}      & 0-Shot & 0.413          & 0.384          & 0.406          & 0.411          & 0.455          & 0.188          & 0.193          & 0.228          & 0.240          & 0.432          \\
		                              & 2-Shot & 0.411          & 0.389          & 0.421          & 0.443          & \textbf{0.471} & 0.275          & \textbf{0.267} & 0.287          & 0.276          & \textbf{0.459} \\
		                              & 0-CoT  & 0.340          & 0.367          & 0.395          & 0.422          & 0.442          & 0.202          & 0.217          & 0.198          & 0.245          & 0.405          \\
		                              & 2-CoT  & 0.396          & 0.382          & 0.399          & 0.443          & 0.447          & 0.254          & 0.260          & 0.260          & 0.275          & 0.437          \\
		\hdashline
		\multirow{3}{*}{GPT-4}        & 0-Shot & \textbf{0.435} & 0.390          & 0.420          & \textbf{0.456} & 0.442          & 0.134          & 0.084          & 0.126          & 0.112          & 0.401          \\
		                              & 2-Shot & 0.421          & \textbf{0.400} & \textbf{0.440} & 0.431          & 0.460          & 0.269          & 0.138          & \textbf{0.288} & 0.210          & 0.436          \\
		                              & 0-CoT  & 0.412          & 0.395          & 0.410          & 0.451          & 0.441          & 0.161          & 0.123          & 0.184          & 0.154          & 0.404          \\
		                              & 2-CoT  & 0.412          & 0.373          & 0.431          & 0.421          & 0.439          & 0.261          & 0.240          & 0.273          & 0.228          & 0.413          \\
		\hdashline
		\multirow{3}{*}{Mixtral-8x7B} & 0-Shot & 0.329          & 0.328          & 0.340          & 0.367          & 0.343          & 0.173          & 0.168          & 0.206          & 0.179          & 0.287          \\
		                              & 2-Shot & 0.326          & 0.312          & 0.327          & 0.369          & 0.385          & \textbf{0.288} & 0.259          & 0.283          & \textbf{0.279} & 0.427          \\
		                              & 0-CoT  & 0.297          & 0.327          & 0.317          & 0.351          & 0.343          & 0.170          & 0.164          & 0.202          & 0.206          & 0.275          \\
		                              & 2-CoT  & 0.386          & 0.349          & 0.383          & 0.417          & 0.396          & 0.286          & 0.264          & \textbf{0.288} & 0.293          & 0.436          \\
		\hdashline
		Average                       &        & 0.382          & 0.366          & 0.391          & 0.415          & 0.426          & 0.222          & 0.198          & 0.235          & 0.225          & 0.401          \\
		\bottomrule
	\end{tabular}
	\vspace{0.5em}
	\caption{Evaluation results of the five benchmark tasks: Idea Generation (``Idea''), Method Recommendation (``Method''), Outcome Prediction (``Outcome''), Future Work Recommendation (``Future'') and Title Prediction (``Title''). 0-CoT stands for zero-shot CoT and 2-CoT stands for two-shot CoT. The models with the best performance are \textbf{bolded}.}
	\label{tab:experiment_combined_results}. 
\end{table}

The results of the benchmark experiments, presented in Table~\ref{tab:experiment_combined_results}, offer several interesting observations:
\begin{itemize}
    \item \textbf{Task Complexity:} On average, LLMs achieve the highest performance on the title prediction task, likely because this task requires minimal extrapolation from the provided information. Among the workflow prediction tasks, outcome prediction and future work recommendation are seemingly easier, although still presenting challenges. Outcome prediction often shows higher performance, likely because published work more often reports positive results, making it somewhat predictable. Interestingly, future work prediction tends to show a better performance than key idea prediction, even though these two tasks are more homogeneous in nature: both extrapolating new directions from the status quo.
    This is likely because many papers include only a cursory discussion of ``projected'' future directions, which tends to be more straightforward and predictable than ``real'' follow-up research that would lead to a future publication. Idea generation and method recommendation are inherently harder, since they require both highly specialized knowledge in the domain and strong innovation capability.
    \item \textbf{Model Performance:} Among the models tested, GPT-4 consistently outperforms GPT-3.5 and Mixtral-8x7B when evaluated by BLEURT, which measures semantic-based similarity. The few-shot prompting method significantly enhances model performance over other methods by helping the models to understand the narrative structure and focus required for the tasks. In contrast, adding CoT to zero-shot prompts or using few-shot CoT does not yield significant improvements, indicating that the complexity of the scientific innovation tasks might exceed the reasoning capabilities of the off-the-shelf LLMs, which are trained on very differen tasks.
\end{itemize}

Overall, our benchmark experiments offer a demonstration of the unique utilities of the MASSW dataset, and they highlight the complexities and nuances of integrating AI models into scientific workflows. The demonstrated tasks are by no means the complete set, and the benchmarked models are by no means the only ones available, nor the best ones. With this new dataset, 
additional tasks of AI-assisted scientific discovery can be designed, and additional AI/machine learning models can be tested and optimized. Readers may consider to use part of this dataset to test various instructing/prompting strategies, fine-tune LLMs for scientific reasoning, or to implement retrieval-augmented solutions.
Furthermore, the current evaluation metrics, which primarily assess semantic and lexical similarity, may not adequately reflect the nuances of the tasks. AI model could have generated meaningful ideas or research methods that are completely different from what's reported in the original paper. More sophisticated evaluation procedures or metrics could be advantageous given the rich and structured information in MASSW.

\section{Related Work}
\label{sec:related_work}
\paragraph{Aspect-based document summarization}

Aspect-based document summarization generates summaries focused on specific document aspects rather than providing a general overview. These aspects may be predefined \citep{E2023-vd, Santosh2024-co, Frermann2019-zn, Takeshita2024-sv, Fisas2015-ai} or dynamically determined based on content \citep{Amar2023-th, Xu2011-fw, Coavoux2019-ob, noauthor_undated-mf, Hayashi2021-dy}. In our case, the aspects are predefined with domain knowledge, identifying five major aspects inherent in scientific workflows.
Aspect-based summarization has been widely applied across various domains. For instance, in the legal domain, \citep{Santosh2024-co} developed a challenging dataset for summarizing legal case decisions. In the context of online shopping, \citep{Coavoux2019-ob} and \citep{Xu2011-fw} have explored dynamic generation of multiple aspect-based summaries for online reviews. Related to our work, \citep{mei2008generating}, \citep{Takeshita2024-sv} and \citep{Fisas2015-ai} have created annotated datasets for summarizing publications in information retrieval, natural language processing, and computer graphics, respectively.
Our approach differs significantly in scope and objective from these studies. Their goal is to establish benchmarks for evaluating models' summarization capabilities, and therefore their end product is usually a limited set of human-annotated examples.  Our aim is to develop a comprehensive, large-scale dataset of scientific workflows, where LLMs, after validation, are used as a proxy for human experts to generate the dataset. Our purpose of creating this dataset is to support extensive downstream tasks related to AI-assisted scientific innovation (such as key idea generation), by including a much larger volume of scientific publications and tailoring the definition of aspects so that they are closely tied to the exploration of scientific workflows.

\paragraph{Scientific workflow automation}

With the rise of LLMs and autonomous agents, many studies investigate the potential of using LLM agents to engage with certain components of scientific workflows, traditionally managed solely by human researchers. \citep{Huang2024-uk} proposes using domain-knowledge-augmented LLM agents to automate and enhance the design of CRISPR-based gene-editing experiments. \citep{Liu2024-up} finds that GPT-4 is useful in converting experimental workflow ideations into executable code on microscope APIs. \citep{Boiko2023-su} shows that an AI system driven by GPT-4 can autonomously design, plan, and execute multiple complex experiments in chemical syntheses. \citep{Agarwal2024-uv} provides an LLM-based toolkit for reviewing scientific literature on a given topic, utilizing retrieval augmented generation to access the latest research. \citep{Procko2023-df} uses LLMs to enhance scientific writing by creating a taxonomy of paper structures, thereby improving efficiency in the academic publishing pipeline. While these preliminary studies focus on particular domains and individual use cases, a significant gap remains in systematically measuring the effectiveness of LLM agents in planning and navigating scientific workflows in general. Our work addresses this gap by introducing multiple new benchmark tasks that assess the capabilities of LLM agents across various critical stages of the scientific process.

\section{Discussion}
\label{sec:conclusion}
\paragraph{Limitations and Future Work}
A limitation of MASSW is that it derives scientific workflows solely from the titles and abstracts of publications, primarily due to the costs associated with accessing and processing large volumes of full papers. It may be potentially more effective to summarize the workflows from multiple sections, given full-text access to these papers. The current scope of this dataset is limited to AI-related, computer science conferences. The defined aspects, although already general, may not always apply to other domains. Addressing these limitations and expanding the practice to cover the literature in other fields will be the focus of the next iterations of the dataset.

\paragraph{Potential Societal Impacts}
We anticipate that MASSW would unleash the great potential of building AI tools to optimize scientific workflows and therefore accelerate the progress in AI for Science. We are, however, aware of two potentially negative impacts. First, by the selection of top-tier AI-related conferences, MASSW might introduce biases to the downstream AI copilots, potentially diminishing the influence of other venues
and limiting the diversity of research topics and methodologies considered. This may be addressed by iterating and expanding the scope of MASSW. Second, the reliance on AI-generated summaries and recommendations could lead researchers to depend on these tools, reducing their engagement with the original papers, and therefore overlook the nuances documented in the literature. This remains an open question for human-AI collaboration.

\acksection
The authors would like to extend their sincere appreciation to Shixuan Liu and Suliang Jin\footnote{Other three annotators (Xingjian Zhang, Ziyang Xiong, and Jin Huang) are authors of this paper.} for their contributions as annotators. This work is funded in part by the LG AI Research Partnership with the University of Michigan.

\newpage
\bibliography{reference.bib}
\bibliographystyle{plainnat}

\appendix
\newpage
\section{Checklist}
\section*{Checklist}

\begin{enumerate}

\item For all authors...
\begin{enumerate}
  \item Do the main claims made in the abstract and introduction accurately reflect the paper's contributions and scope?
    \answerYes{The main claims made in the abstract and introduction accurately reflect the paper's contributions and scope.}
  \item Did you describe the limitations of your work?
    \answerYes{See Section~\ref{sec:conclusion}}
  \item Did you discuss any potential negative societal impacts of your work?
    \answerYes{See Section~\ref{sec:conclusion}}
  \item Have you read the ethics review guidelines and ensured that your paper conforms to them?
    \answerYes{All authors conform to the ethics review guidelines.}
\end{enumerate}

\item If you are including theoretical results...
\begin{enumerate}
  \item Did you state the full set of assumptions of all theoretical results?
    \answerNA{We did not include theoretical results.}
	\item Did you include complete proofs of all theoretical results?
    \answerNA{We did not include theoretical results.}
\end{enumerate}

\item If you ran experiments (e.g. for benchmarks)...
\begin{enumerate}
  \item Did you include the code, data, and instructions needed to reproduce the main experimental results (either in the supplemental material or as a URL)?
    \answerYes{See Abstract and Appendix~\ref{app:url}}
  \item Did you specify all the training details (e.g., data splits, hyperparameters, how they were chosen)?
    \answerYes{See Appendix~\ref{sec:experiment-details}}
	\item Did you report error bars (e.g., with respect to the random seed after running experiments multiple times)?
    \answerNA{We did not present any models that require training (fine-tuning) in the downstream tasks. We believed the randomness from pre-trained LLM inferences are minimal (with a zero temperature) therefore we did not report error bars.}
	\item Did you include the total amount of compute and the type of resources used (e.g., type of GPUs, internal cluster, or cloud provider)?
    \answerYes{See Appendix~\ref{sec:experiment-details}}
\end{enumerate}

\item If you are using existing assets (e.g., code, data, models) or curating/releasing new assets...
\begin{enumerate}
  \item If your work uses existing assets, did you cite the creators?
    \answerYes{See Section~\ref{sec:dataset_overview}}
  \item Did you mention the license of the assets?
    \answerYes{See Section~\ref{sec:data-curation}}
  \item Did you include any new assets either in the supplemental material or as a URL?
    \answerNo{}
  \item Did you discuss whether and how consent was obtained from people whose data you're using/curating?
    \answerNo{The existing dataset (OAG) that we used is publicly available under the ODC-BY license. }
  \item Did you discuss whether the data you are using/curating contains personally identifiable information or offensive content?
    \answerNA{}
\end{enumerate}

\item If you used crowdsourcing or conducted research with human subjects...
\begin{enumerate}
  \item Did you include the full text of instructions given to participants and screenshots, if applicable?
    \answerNA{We did not use crowdsourcing or conduct research with human subjects.}
  \item Did you describe any potential participant risks, with links to Institutional Review Board (IRB) approvals, if applicable?
    \answerNA{Ditto.}
  \item Did you include the estimated hourly wage paid to participants and the total amount spent on participant compensation?
    \answerNA{Ditto.}
\end{enumerate}

\end{enumerate}

\newpage
\section{Supplementary Material}

\subsection{Data Curation}
\label{app:data-curation}

To build this initial version of the MASSW dataset, we focus on Computer Science publications from 17 top-tier conferences listed in \href{CSRankings.org}{CSRankings.org}, which we identify as relevant to the broader field of AI. Below we list the conferences included in MASSW. 

\begin{itemize}
    \item Artificial Intelligence: AAAI, IJCAI;
    \item Computer Vision: CVPR, ECCV, ICCV;
    \item Machine Learning: ICLR, ICML, NeurIPS, KDD;
    \item Natural Language Processing: ACL, EMNLP, NAACL;
    \item The Web \& Information Retrieval: SIGIR, WWW;
    \item Databases: SIGMOD, VLDB; 
    \item Interdisciplinary Areas: CHI.
\end{itemize}

We access the publications through Open Academic Graph (OAG)\footnote{The OAG dataset is publicly released under the ODC-BY license.}, a linked graph database for academic entities including publications, venues, affiliations, and authors
\citep{zhang2022oag,zhang2019oag}. 
For publications before the year 2020, we access the data through OAG v2.1\footnote{OAG v2.1: \url{https://old.aminer.cn/oag-2-1/oag-2-1}.}, which is generated in 2020 and contains publications as early as 1969. For publications in and after 2020, we access the data through OAG v3.1\footnote{OAG v3.1: \url{https://open.aminer.cn/open/article?id=65bf053091c938e5025a31e2}.}, which is generated in Feb, 2024 and contains publications from 2000 to 2024. 

In total, 191,055 papers are collected that span from 1969 to 2024, among which, 152,027 contain both a title and an abstract.

\subsection{Aspect Summarization}
\label{app:aspect-summ}
\label{sec:example-aspects}

We use OpenAI GPT-4 (snapshot \texttt{gpt-4-0613}\footnote{OpenAI GPT-4 models: \url{https://platform.openai.com/docs/models/gpt-4-turbo-and-gpt-4}. }) to summarize the core aspects of each collected publication. Here we provide the prompt we used for automated summarization:

\begin{prompt}{Aspect Summarization Prompt}
\textbf{System message:}

Instructions \\
You are an expert in computer science.
Your task is to summarize the following five aspects of the papers given the
definitions below. \\

Definitions of Aspects\\
Context\\
- The status quo of related literature or reality which motivated this study. This could normally be a problem, a research question, or a research gap that has not been successfully addressed by previous work. \\
- Anything happened before this study.\\
Key Idea\\
- The main intellectual merit of this paper, often in comparison to the
context. This could normally be a novel idea or solution proposed in this paper that distinguishes it from what's already done in literature. \\
- Proposed in this study. \\
\end{prompt}

\begin{prompt}{Aspect Summarization Prompt (Cont'd)}
Method (Validation Methodology)\\
- The specific experiment or proof that investigates and validates the key idea.\\
- CS papers often refer "Method" as algorithm or model, but our definition here is **different**.\\
- Performed in this study.\\
Outcome\\
- The factual statement about the study output. This could be the experiment
results and any other measurable outcome that has occurred. It marks whether the key hypothesis is testified or not.\\
- Produced in this study.\\
Future Impact\\
- The impact of the work on the field explicitly anticipated by the authors, and potential further research explicitly identified by the author that may improve or extend this study.\\

Notes\\
- If an aspect is NOT mentioned in the abstract, mark it as "N/A" (not applicable). DO NOT come up with your own interpretation. \\
- Each aspect should be summarized in 1-2 sentences in most cases. \\
- Each aspect should be self-contained and should not contain references including other aspects (cross-reference). \\
- Including specific names of proposed models, datasets, etc., in the summary is acceptable.\\
- If the problem definition is novel (e.g., proposing a new task), classify it as a Key Idea.\\
- Non-measurable outcomes should be categorized as Future Impact.\\
- Impacts that have already occurred should be considered as Outcome.\\
- A new observation that motivates the proposal of a key idea should be classified under Key Idea.\\
- Future Impact should not account for real impacts, such as the number of
citations a paper has received.\\

Response Format\\
The response should be a JSON object in the following format:\\
```json\\
\{\\
    "Context": "...",\\
    "Key Idea": "...",\\
    "Method": "...",\\
    "Outcome": "...",\\
    "Future Impact": "..."\\
\}\\
```\\

\textbf{User message:}\\
Title: Attention Is All You Need\\
Abstract: The dominant sequence transduction models are based on complex recurrent or convolutional neural networks in an encoder-decoder configuration. The best performing models also connect the encoder and decoder through an attention mechanism. We propose a new simple network architecture, the Transformer, based solely on attention mechanisms, dispensing with recurrence and convolutions entirely. Experiments on two machine translation tasks show these models to be superior in quality while being more parallelizable and requiring significantly less time to train. Our model achieves 28.4 BLEU on the WMT 2014 English-to-German translation task, improving over the existing best results, including ensembles by over 2 BLEU. On the WMT 2014 English-to-French translation task, our model establishes a new single-model state-of-the-art BLEU score of 41.8 after training for 3.5 days on eight GPUs, a small fraction of the training costs of the best models from the literature. We show that the Transformer generalizes well to other tasks by applying it successfully to English constituency parsing both with large and limited training data.
\end{prompt}

\begin{prompt}{Aspect Summarization Prompt (Cont'd)}
\textbf{Assistant message:}\\
\{\\
  "Context": "The dominant sequence transduction models are based on complex recurrent or convolutional neural networks in an encoder-decoder configuration. The best performing sequence transduction models connect the encoder and decoder through an attention mechanism.",\\
  "Key Idea": "The authors propose a simple network architecture called Transformer based solely on attention mechanisms and dispenses with recurrence and convolutions.",\\
  "Method": "The authors perform experiments on the WMT 2014 English-to-German and English-to-French translation task. The authors apply the proposed model to English constituency parsing both with large and limited training data.",\\
  "Outcome": "The proposed model achieves a BLEU score of 28.4 on the WMT 2014 English-to-French translation task. The proposed model achieves a BLEU score of 41.8 on the WMT 2014 English-to-German translation task after training for 3.5 days on 8 GPUs.",\\
  "Future Impact": "N/A"\\
\}\\

\textbf{User prompt:}\\
Title: [The title of the publication to be summarized]\\
Abstract: [The abstract of the publication to be summarized]
\end{prompt}

In the prompt, the first user message and the assistant message provide a one-shot example to the LLM. During automated summarization, GPT-4 failed to summarize 12 publications due to safety filtering. In total, we obtained core aspects from 152,015 publications. 

\begin{table}[h]
    \centering
    \small
    \scalebox{0.8}{
    \begin{tabular}{m{1.25\textwidth}}
        \toprule
        \textbf{Title}: Attention Is All You Need \\
        \textbf{Abstract}: The dominant sequence transduction models are based on complex recurrent or convolutional neural networks in an encoder-decoder configuration. The best performing models also connect the encoder and decoder through an attention mechanism. We propose a new simple network architecture, the Transformer, based solely on attention mechanisms, dispensing with recurrence and convolutions entirely. Experiments on two machine translation tasks show these models to be superior in quality while being more parallelizable and requiring significantly less time to train. Our model achieves 28.4 BLEU on the WMT 2014 English-to-German translation task, improving over the existing best results, including ensembles by over 2 BLEU. On the WMT 2014 English-to-French translation task, our model establishes a new single-model state-of-the-art BLEU score of 41.8 after training for 3.5 days on eight GPUs, a small fraction of the training costs of the best models from the literature. We show that the Transformer generalizes well to other tasks by applying it successfully to English constituency parsing both with large and limited training data. \\
        \textbf{LLM Summarized Core Aspects:}\\
        \textbf{Context}: Prior to this work, sequence transduction models were dominated by complex recurrent and convolutional neural networks in an encoder-decoder configuration, with optimal results found through the use of attention mechanisms connecting the encoder and decoder. \\
        \textbf{Key Idea}: The paper introduces a new network architecture, the Transformer, that exclusively relies on an attention mechanism, foregoing recurrence and convolutions entirely. \\
        \textbf{Method}: Experiments were conducted on two machine translation tasks to validate the proposed network architecture, comparing its quality, parallelism, and training speed against the existing models. \\
        \textbf{Outcome}: The Transformer model outperformed existing models in terms of quality and training efficiency, achieving a BLEU score of 27.5 on English to German translation, which improved over the current best ensemble result by over 1 BLEU, and a BLEU score of 41.1 on English to French translation, surpassing the previous single state-of-the-art model by 0.7 BLEU. \\
        \textbf{Projected Impact}: N/A \\
        \midrule
        \textbf{Title}: Deep Residual Learning for Image Recognition
 \\
        \textbf{Abstract}: Deeper neural networks are more difficult to train. We present a residual learning framework to ease the training of networks that are substantially deeper than those used previously. We explicitly reformulate the layers as learning residual functions with reference to the layer inputs, instead of learning unreferenced functions. We provide comprehensive empirical evidence showing that these residual networks are easier to optimize, and can gain accuracy from considerably increased depth. On the ImageNet dataset we evaluate residual nets with a depth of up to 152 layers---8x deeper than VGG nets but still having lower complexity. An ensemble of these residual nets achieves 3.57\% error on the ImageNet test set. This result won the 1st place on the ILSVRC 2015 classification task. We also present analysis on CIFAR-10 with 100 and 1000 layers. The depth of representations is of central importance for many visual recognition tasks. Solely due to our extremely deep representations, we obtain a 28\% relative improvement on the COCO object detection dataset. Deep residual nets are foundations of our submissions to ILSVRC \& COCO 2015 competitions, where we also won the 1st places on the tasks of ImageNet detection, ImageNet localization, COCO detection, and COCO segmentation. \\
        \textbf{LLM Summarized Core Aspects:}\\
        \textbf{Context}: The deeper neural networks become, the more difficult they are to train, creating a barrier for network depth improvement.\\
        \textbf{Key Idea}: The authors develop a new residual learning framework that facilitates the training of substantially deeper networks, by reformulating the layers to learn residual functions with reference to the layer inputs.\\
        \textbf{Method}: Empirical evidence is provided through experimentation on the ImageNet dataset with residual nets of up to 152 layers, and the analysis on CIFAR-10 with 100 and 1000 layers. \\
        \textbf{Outcome}: Residual networks lead to ease of optimization and gain in accuracy with increased depth. An ensemble of these residual nets achieved a 3.57\% error on the ImageNet test set, winning the 1st place on the ILSVRC 2015 classification task. There was a 28\% relative improvement on the COCO object detection dataset due to the extremely deep representations. \\
        \textbf{Projected Impact}: The paper paves the way for more elaborated deep learning models, as it demonstrates the potential for creating deeper nets that are easier to optimize. The success in several competitions highlights their potential for further applications in image detection and localization tasks. \\
        \midrule
        \textbf{Title}: ImageNet: A large-scale hierarchical image database \\
        \textbf{Abstract}: The explosion of image data on the Internet has the potential to foster more sophisticated and robust models and algorithms to index, retrieve, organize and interact with images and multimedia data. But exactly how such data can be harnessed and organized remains a critical problem. We introduce here a new database called “ImageNet”, a large-scale ontology of images built upon the backbone of the WordNet structure. ImageNet aims to populate the majority of the 80,000 synsets of WordNet with an average of 500–1000 clean and full resolution images. This will result in tens of millions of annotated images organized by the semantic hierarchy of WordNet. This paper offers a detailed analysis of ImageNet in its current state: 12 subtrees with 5247 synsets and 3.2 million images in total. We show that ImageNet is much larger in scale and diversity and much more accurate than the current image datasets. Constructing such a large-scale database is a challenging task. We describe the data collection scheme with Amazon Mechanical Turk. Lastly, we illustrate the usefulness of ImageNet through three simple applications in object recognition, image classification and automatic object clustering. We hope that the scale, accuracy, diversity and hierarchical structure of ImageNet can offer unparalleled opportunities to researchers in the computer vision community and beyond. \\
        \textbf{LLM Summarized Core Aspects:}\\
        \textbf{Context}: The exponential increase of image data available on the internet provides potential for more sophisticated and robust image-based models and algorithms. However, harnessing and organizing such data effectively poses a significant challenge. \\
        \textbf{Key Idea}: The authors introduce a new database, ImageNet, a large-scale ontology of images built upon the structure of WordNet. The goal of ImageNet is to populate the majority of the 80,000 WordNet synsets with an average of 500-1000 clean, full-resolution images. \\
        \textbf{Method}: The authors create ImageNet using a data collection scheme with Amazon Mechanical Turk, and analyse and compare ImageNet in its current state to other existing datasets. They illustrate its usefulness through object recognition, image classification, and automatic object clustering. \\
        \textbf{Outcome}: ImageNet, in its current state, has 12 subtrees with 5247 synsets and 3.2 million images. It is found to be larger and more diverse than the existing image datasets and offers more accuracy. \\
        \textbf{Projected Impact}: The authors anticipate that the scale, accuracy, diversity, and hierarchical structure of ImageNet will offer unparalleled research opportunities to the computer vision community and beyond. \\

        \bottomrule
    \end{tabular}
    }
    \caption{Examples of summarized core aspects in MASSW. }
    \label{tab:aspect-examples}
\end{table}

\subsection{Dataset Visualization}
\label{app:data-vis}

\begin{figure}[!ht]
\centering
\begin{subfigure}{0.49\textwidth}
    \includegraphics[width=\textwidth]{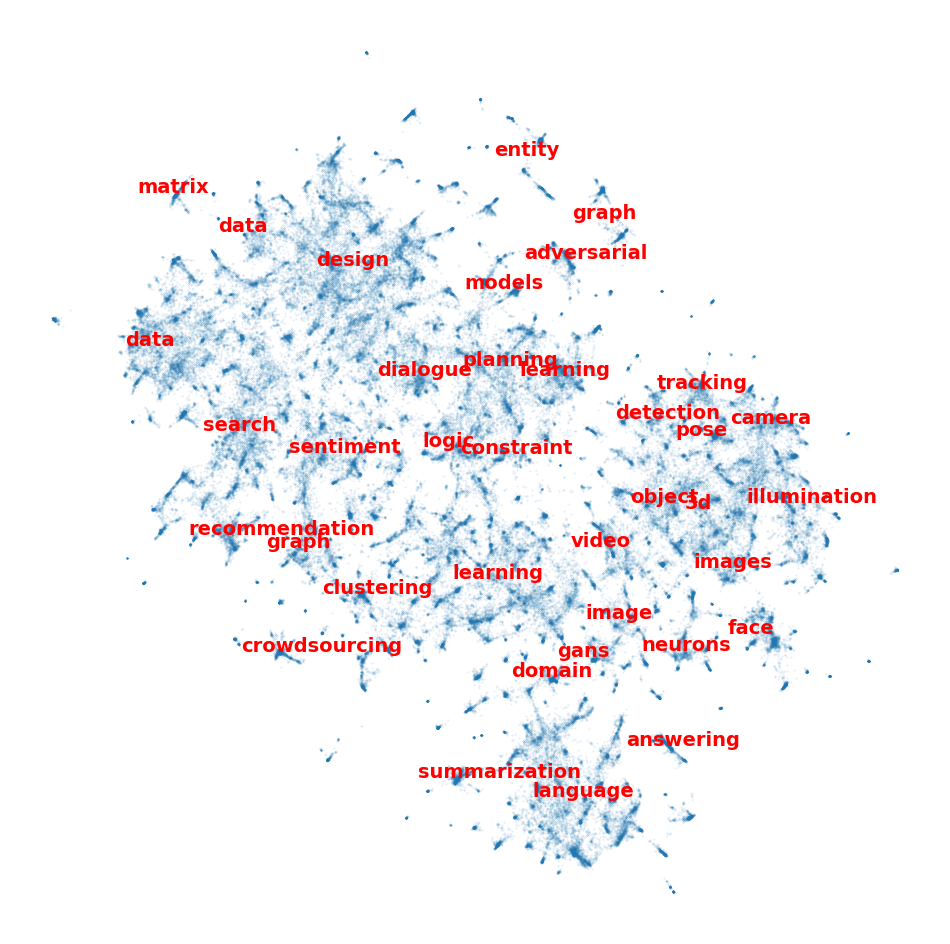}
    \caption{Context.}
    \label{fig:largevis-context}
\end{subfigure}
\hfill
\begin{subfigure}{0.49\textwidth}
    \includegraphics[width=\textwidth]{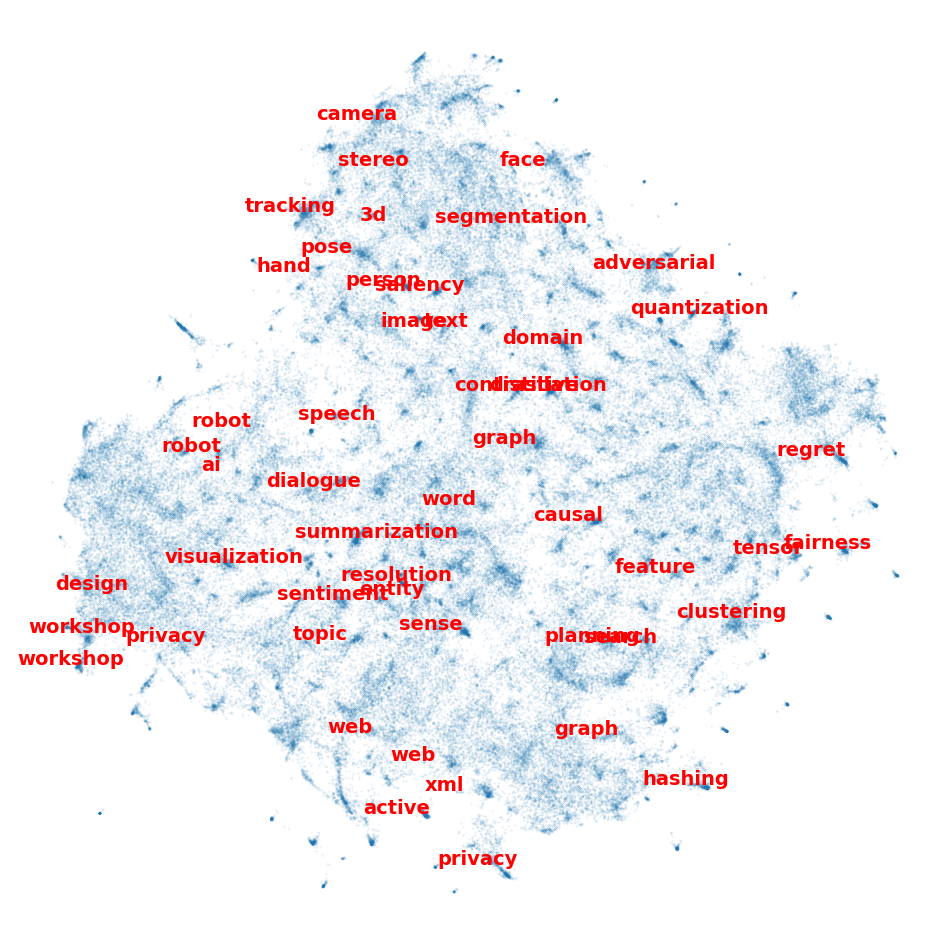}
    \caption{Key Idea.}
    \label{fig:largevis-key_idea}
\end{subfigure}
\hfill
\begin{subfigure}{0.49\textwidth}
    \includegraphics[width=\textwidth]{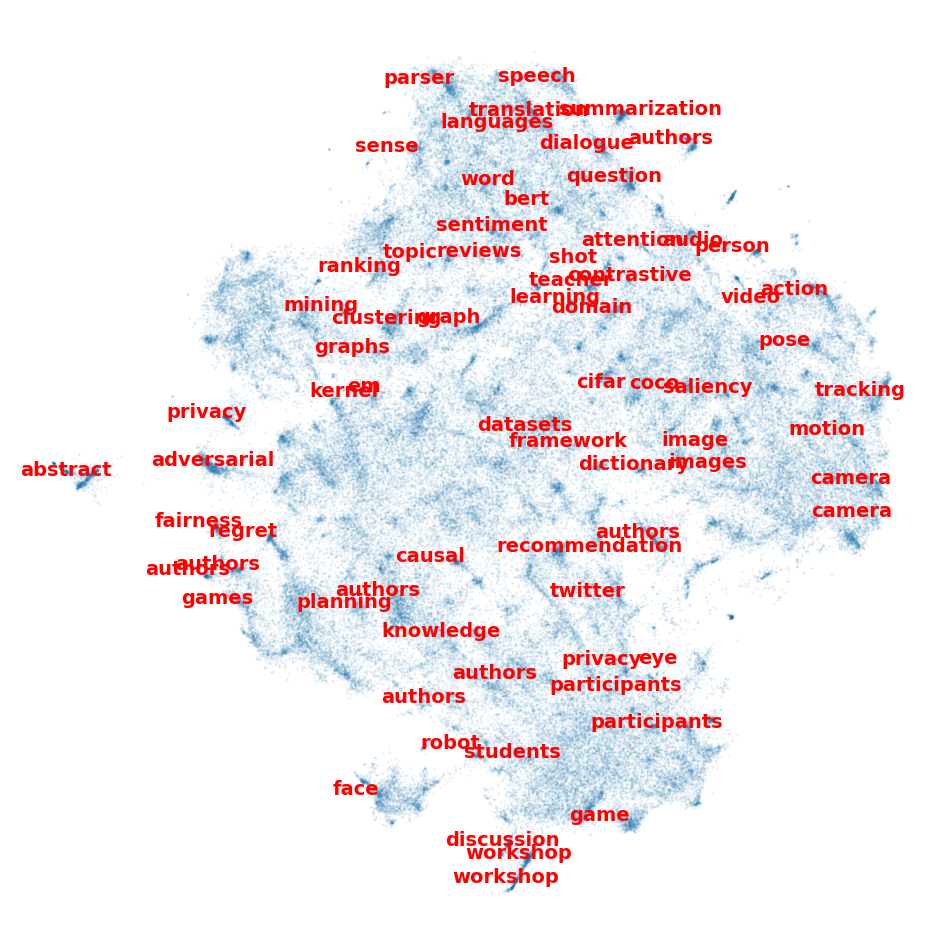}
    \caption{Method.}
    \label{fig:largevis-method}
\end{subfigure}
\hfill
\begin{subfigure}{0.49\textwidth}
    \includegraphics[width=\textwidth]{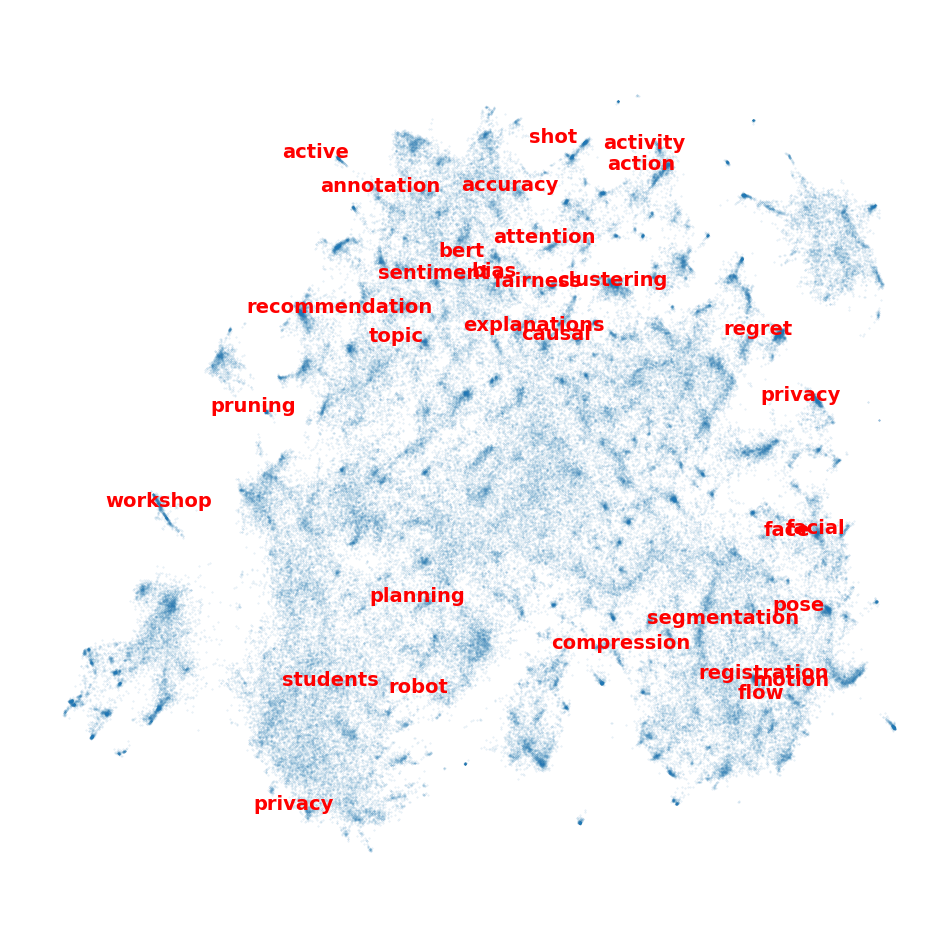}
    \caption{Outcome.}
    \label{fig:largevis-outcome}
\end{subfigure}
\hfill
\begin{subfigure}{0.49\textwidth}
    \includegraphics[width=\textwidth]{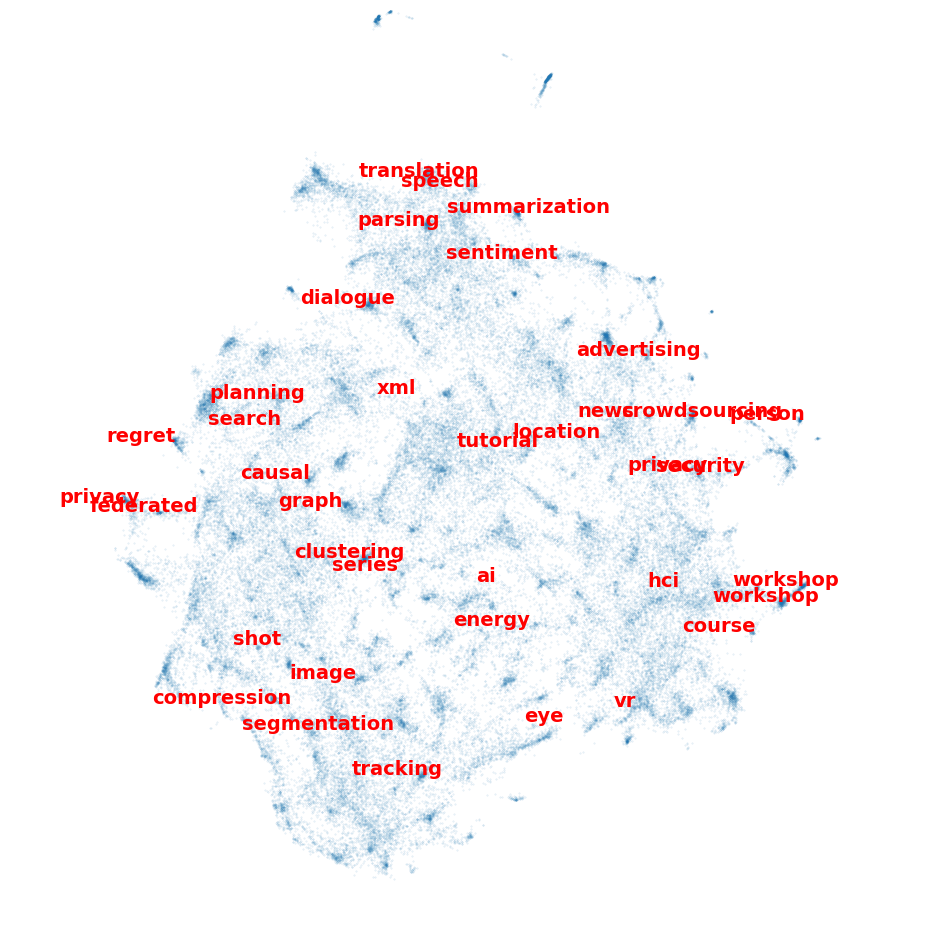}
    \caption{Projected Impact.}
    \label{fig:largevis-future_impact}
\end{subfigure}
\caption{Visualizations of MASSW aspects. }
\label{fig:largevis-aspects}
\end{figure}

Figure \ref{fig:largevis-aspects} visualizes the summarized core aspects from publications in multiple views. 
For each subplot, we first embed aspect summaries of publications with OpenAI Ada 3\footnote{OpenAI Embedding Models: \url{https://platform.openai.com/docs/models/embeddings}. }, then conduct dimension reduction with LargeVis \citep{tang2016visualizing}. Clusters identified by HDBSCAN are annotated with TF-IDF top words. 

\subsection{Human Annotation Process}
\label{sec:annotation}
\subsubsection{Overview}
We engage five student researchers from the University of Michigan, each with a verified background in AI, to perform the annotations. To minimize individual bias, each paper is independently annotated by two different researchers. The annotations are carried out using the Potato annotation platform~\citep{Pei2022-ec}.
\subsubsection{Codebook}

\paragraph{Task Description}
Our task is to construct a dataset for multi-aspect summarization of scientific papers. Our papers of interest are from top computer science conferences. For each paper, the aspects of interest include the following: (Same content in Table~\ref{tab:study_framework})

Your task is to write summarizations of these five aspects for each paper assigned to you. We have the following requirements for this task:
\begin{itemize}
    \item Read the content thoroughly before writing your summaries.
    \item Write a short summary for each aspect (1-2 sentences in most cases).
    \item Each aspect should be self-contained and should not contain references including other aspects (cross reference).
    \item Only consider the abstract section and title as the input.
\end{itemize}

\paragraph{FAQ}

\begin{itemize}
    \item \textbf{Q:} Is it fine to include the specific name of the proposed model/dataset/etc in the summary? \\
    \textbf{A:} Yes, it is fine to include them.
    \item \textbf{Q:} If the problem definition is novel (i.e. proposing a new task), should it be a key idea or context? \\
    \textbf{A:} Key idea.
    \item \textbf{Q:} If the concept is not mentioned at all in the abstract, what should I do? \\
    \textbf{A:} Mark it as “N/A” (not applicable).
    \item \textbf{Q:} If the author claims a non-measurable outcome, should it be considered as an Outcome or Future Impact? \\
    \textbf{A:} Future Impact.
    \item \textbf{Q:} If the author mentions an impact that has happened (e.g. the first work to …), should it be considered as an Outcome or Future Impact? \\
    \textbf{A:} Outcome.
    \item \textbf{Q:} If the author mentions a new observation that motivates them to propose the key idea, should it be considered as context or key idea? \\
    \textbf{A:} Key idea.
    \item \textbf{Q:} Should future impact consider its real impact? For example, a paper gains a lot of citations. \\
    \textbf{A:} Future Impact should not consider other papers.
\end{itemize}

\subsection{Implementation Details of Semantic-Based Evaluation Metrics}
\label{sec:semantic-metrics}
\begin{itemize}
    \item \textit{Cosine Similarity}: We compute the cosine similarity between sentence embeddings generated by \texttt{intfloat/multilingual-e5-large-instruct} from HuggingFace.
    \item \textit{BLEURT}: We use the pre-trained checkpoint \texttt{BLEURT-20-D12}.
    \item \textit{BERTScore}: We use the pre-trained checkpoint from HuggingFace \url{https://huggingface.co/spaces/evaluate-metric/bertscore}.
\end{itemize}
We follow the implementation of ROUGE to select the maximum score when there are multiple references.

\subsection{Examples of Texts for Different Similarity Levels}
\label{sec:example-similarity}
We provide two examples of texts to illustrate how the evaluation metrics could be interpreted. The evaluation results can be found at Table~\ref{tab:similarity_scores1} and \ref{tab:similarity_scores2}.
\begin{itemize}
    \item \textbf{Reference 1:} InstructGPT, even with 100x fewer parameters, is preferred over GPT-3 in human evaluations. It shows improvements in truthfulness and reductions in toxic outputs with minimal performance regressions on public NLP datasets.
    \item \textbf{Example 1a:} InstructGPT, despite having 100x fewer parameters, is preferred over the larger GPT-3 according to human evaluations, demonstrating better truthfulness and fewer toxic outputs with only minimal regressions in performance on public NLP benchmarks.
    \item \textbf{Example 1b:} Human evaluations favor the 1.3B parameter InstructGPT model over the 175B GPT-3 model, even though it has significantly fewer parameters. It also shows enhanced truthfulness and reduced generation of toxic content, with negligible declines in performance across standard NLP datasets.
    \item \textbf{Example 1c:} In human assessments, the smaller InstructGPT model, which has far fewer parameters, outperforms GPT-3, showing not only increased accuracy but also less toxic output, with only slight performance downturns on widely recognized NLP tests.
    \item \textbf{Example 1d:} This paper explores the enhancement of language model alignment with human intent through fine-tuning methods using labeler feedback and reinforcement learning, resulting in a smaller, more efficient model that surpasses a much larger baseline in both user satisfaction and safety metrics.
    \item \textbf{Example 1e:} Effective communication is not about speaking more; it’s about achieving more with fewer words.
\end{itemize}

\begin{table}[!h]
\centering
\begin{tabular}{cccccc}
\toprule
\textbf{Example} & \textbf{CS} & \textbf{BLEURT} & \textbf{BS} & \textbf{BLEU} & \textbf{ROUGE-1} \\ \midrule
1a               & 0.9500          & 0.7185          & 0.9589                & 0.2753        & 0.6970           \\ 
1b               & 0.9366          & 0.6202          & 0.9188                & 0.0000        & 0.5135           \\ 
1c               & 0.9326          & 0.5572          & 0.9109                & 0.0000        & 0.3582           \\ 
1d               & 0.8384          & 0.3119          & 0.8504                & 0.0000        & 0.1351           \\ 
1e               & 0.7594          & 0.1953          & 0.8396                & 0.0000        & 0.1702           \\ \bottomrule
\end{tabular}
\vspace{1em}
\caption{Evaluation of similarity between examples and Reference 1 using various metrics.}
\label{tab:similarity_scores1}
\end{table}

\begin{itemize}
    \item \textbf{Reference 2:} The dominant sequence transduction models are based on complex recurrent or convolutional neural networks in an encoder-decoder configuration. The best performing sequence transduction models connect the encoder and decoder through an attention mechanism.
    \item \textbf{Example 2a:} The leading sequence transduction models utilize complex recurrent or convolutional neural networks in an encoder-decoder framework, with the most effective models incorporating an attention mechanism between the encoder and decoder.
    \item \textbf{Example 2b:} Traditional sequence transduction models rely on sophisticated recurrent or convolutional neural networks arranged in an encoder-decoder setup, where top-performing models are distinguished by the use of an attention mechanism linking the encoder and decoder.
    \item \textbf{Example 2c:} Existing high-performing sequence transduction models typically feature either recurrent or convolutional neural networks configured in an encoder-decoder structure, often enhanced with an attention mechanism to improve performance.
    \item \textbf{Example 2d:} The paper introduces the Transformer, a novel network architecture that eschews recurrent and convolutional structures in favor of a design entirely based on attention mechanisms, aiming to enhance parallelizability and reduce training time.
    \item \textbf{Example 2e:} "Simplicity is the ultimate sophistication." - Leonardo da Vinci
\end{itemize}

\begin{table}[!h]
\centering
\begin{tabular}{cccccc}
\toprule
\textbf{Example} & \textbf{CS} & \textbf{BLEURT} & \textbf{BS} & \textbf{BLEU} & \textbf{ROUGE-1} \\ \midrule
2a               & 0.9572          & 0.7256          & 0.9613                & 0.3772        & 0.7077           \\
2b               & 0.9516          & 0.6781          & 0.9494                & 0.2689        & 0.6857           \\
2c               & 0.9381          & 0.5660          & 0.9289                & 0.1927        & 0.5079           \\
2d               & 0.8355          & 0.3598          & 0.8645                & 0.0000        & 0.2687           \\
2e               & 0.7108          & 0.1728          & 0.8095                & 0.0000        & 0.0476           \\ \bottomrule
\end{tabular}
\vspace{1em}
\caption{Evaluation of similarity between examples and Reference 2 using various metrics.}
\label{tab:similarity_scores2}
\end{table}

\subsection{Experiment Details}
\label{sec:experiment-details}

\paragraph{Test set sampling.} In the benchmark section, We use proportionate stratified sampling to construct the test set. According to publication year, we separate the year range into at most 10 strata (i.e. groups). Each group covers approximately the same number of years. The we sample from each strata proportionally to the number of papers in that strata. The number of samples for each venue is 60, which results in 1020 papers in total.

\paragraph{Prompting templates.}

We recall that the model will take in three part of information: 
 (i) definitions of all five aspects, (ii) all necessary aspects for each task, and (iii) a specific task instruction. We include the prompts for all tasks below.

\begin{prompt}{Prompt}
\textbf{System message:}

You are an expert in research tasked with generating detailed prompts for various aspects of academic research papers. Each task involves creating a specific type of prompt based on the provided information. Here are the definitions of each part you will work with:

- Concept

\quad  - Definition
  
\quad  - Relative Time

- Context: The status quo of related literature or reality which motivated this study. This could normally be a problem, a research question, or a research gap that has not been successfully addressed by previous work. This is anything that happened before this study.

- Key Idea: The main intellectual merit of this paper, often in comparison to the context. This could normally be a novel idea or solution proposed in this paper that distinguishes it from what’s already done in literature. This is proposed in this study.

- Method: The specific research method that investigates and validates the key idea. This could be an experimental setup, a theoretical framework, or other necessary methodology to implement and/or evaluate the key idea. This is performed in this study.

- Outcome: The factual statement about the study output. This could be the experiment results and any other measurable outcome that has occurred. It marks whether the key hypothesis is testified or not. This is produced in this study.

- Projected Impact: The author-anticipated impact of the work on the field, and potential further research identified by the author that may improve or extend this study. This is anything being anticipated but has not happened yet.
\end{prompt}

\begin{prompt}{Prompt}
\textbf{Template for idea generation:}

Given the context: '\{context\}', generate key ideas that could advance this area of study. 
\\
\\

\textbf{Template for method recommendation:}

Given the context: '\{context\}' and the key idea: '\{key\_idea\}', recommend the most suitable method to validate this idea. 
\\
\\

\textbf{Template for outcome prediction:}

Based on the context: '\{context\}', the key idea: '\{key\_idea\}', and the recommended method: '\{method\}', predict the potential outcome of this research. 
\\
\\

\textbf{Template for impact prediction:}

Based on the context: '\{context\}', the key idea: '\{key\_idea\}', the method: '\{method\}', and the outcome: '\{outcome\}', suggest projected Impact for this research.
\\
\\

\textbf{Template for title prediction:}

Given the context: '\{context\}', the key idea: '\{key\_idea\}', the method: '\{method\}', the outcome: '\{outcome\}', and the future impact: '\{future\_impact\}', predict the title of this research paper. The title should be concise and reflective of the core aspects.

\end{prompt}

\paragraph{Details about prompting methods.} For the zero shot prediction, model will take in the system prompt and user prompt. For the few shot prompting, we add two fixed round of conversation before the actual user request. The few-shot examples can be found in the code-base under the data folder. For chain of thought prompts, we add the sentence ``Let's think step by step. The final prediction should start after the marker 'Prediction:'.'' at the end of zero-shot prompts. After LLMs produce the output, we extract the content after the word ``Prediction'' as the final prediction.

\paragraph{Details about baseline LLMs.} We use Azure platform for all baseline models. The model id used in this study are \texttt{gpt-35-turbo (0125)}\footnote{\url{https://learn.microsoft.com/en-us/azure/ai-services/openai/concepts/models}}, \texttt{gpt-4 (turbo-2024-04-09)}  and \texttt{mistralai-mixtral-8x7b-instru-7}\footnote{\url{https://learn.microsoft.com/en-us/azure/machine-learning/how-to-deploy-models-mistral?view=azureml-api-2&tabs=mistral-large}}. For the mistral model, as the model does not take in system prompt, we replace it with one round of conversation (user: \texttt{<system prompt>}, assistant: \texttt{I got it. Please give me further instructions!}).

\paragraph{Example model outputs.} We include an example output for all the prompting methods in Table~\ref{tab:example_outputs}.

\begin{table}[ht]
\centering
\scriptsize
\begin{tabular}{m{0.1\textwidth}|m{0.85\textwidth}}
\textbf{Prompt Type} & \textbf{Model Output/Ground Truth} \\ 
\hline
\textbf{Ground Truth} & This paper introduces the concept of co-space applications, which provide opportunities unavailable in separate physical or virtual spaces, and posit that the database community can significantly contribute to this emerging field. \\
\hline
\textbf{0-shot} & **Key Idea 1: Development of a Unified Interaction Framework**

- **Concept:** This key idea proposes creating a unified interaction framework that enables users to interact with both physical and virtual elements through a single interface or set of interaction gestures. This framework integrates sensory feedback mechanisms (e.g., haptic, auditory, visual cues) that are context-aware and adapt based on the user's actions and environmental conditions.

- **Relative Time:** Proposed in this study.

**Key Idea 2: Hybrid Space Adaptation Engine**

- **Concept:** The introduction of a hybrid space adaptation engine designed to dynamically modify the virtual space in response to changes in the physical environment and vice versa. This engine uses real-time data analytics and machine learning algorithms to predict and adapt to user needs and environmental shifts, ensuring a consistent and immersive experience across both domains.

- **Relative Time:** Proposed in this study.

<Some other key ideas.> \\

\hline
\textbf{2-shot} & The researchers propose the development of an intuitive Mixed Reality (MR) interface that leverages advancements in augmented reality (AR) and virtual reality (VR), combined with real-time spatial computing and machine learning algorithms. This interface is designed to enable more seamless and intuitive user control and interaction between physical and virtual spaces, enhancing the co-space experience by making digital manipulations feel as tangible as physical ones. \\ 
\hline
\textbf{CoT} & To address the context effectively, a detailed step-by-step approach will be taken to generate key ideas that could significantly advance the study of co-space environments, where the integration and interaction between physical and virtual spaces present both challenges and opportunities for technological innovation. These key ideas are crafted to facilitate a seamless manipulation of real and digital spaces by users.

\#\#\# Concept:
- **Definition**: Co-space environments are settings in which physical and digital realms are intertwined, allowing for continuous interaction and manipulation across these domains.
- **Relative Time**: This concept is becoming increasingly relevant with the advances in augmented reality (AR), virtual reality (VR), mixed reality (MR), and Internet of Things (IoT) technologies.

\#\#\# Key Ideas:

1. **Development of a Unified Interaction Protocol (UIP)**
   - This protocol would standardize the way in which interactions occur between the physical and virtual components of a co-space environment. The UIP would ensure that devices, regardless of their operational technology (OT) or information technology (IT) origins, can communicate and interact seamlessly in a common language.

2. **Real-time Spatial Mapping and Augmentation Technology**
   - An advanced spatial mapping technology that dynamically captures the physical environment and accurately overlays digital information in real-time. This would not only facilitate seamless user manipulation of digital objects in real spaces but also enhance the realism and interactiveness of the virtual space.

<Some other key ideas.>

\#\#\# Prediction:
- The aforementioned key ideas, if successfully implemented, are poised to revolutionize the co-space environment by breaking down the barriers between the physical and virtual realms. These innovations would not only facilitate seamless manipulation of spaces by users but would also set a new standard for interactive technologies, significantly enhancing user experiences while addressing privacy and security concerns. Moreover, the advancement in such areas could pave the way for novel applications ranging from immersive education and remote collaboration to advanced urban planning and beyond.
 \\ 
 \hline
\textbf{2-shot CoT} & Integrating physical and virtual realms in co-space environments poses technical and usability challenges, necessitating novel solutions for seamless interaction. One way forward is to bridge the gap between these spaces with technologies that can intuitively understand and manipulate both contexts in real-time. 

Prediction: The authors propose an innovative interaction framework that blends Augmented Reality (AR) and Internet of Things (IoT) technologies. This framework enables users to manipulate objects in both the physical and the virtual spaces through natural gestures and commands, leveraging AR for visual integration and IoT devices for physical state changes. \\
\end{tabular}
\caption{Comparison of Zero-shot, Few-shot, and Chain of Thought outputs for a single example by GPT-4.}
\label{tab:example_outputs}
\end{table}

\subsection{Supplementary Tables}
\label{sec:supplementary-tables}

We include the benchmark results for Cosine Similarity (CS), BERTScore (BS) and BLEU in Table~\ref{tab:experiment_results_cs},~\ref{tab:experiment_results_bs},~\ref{tab:experiment_results_bleu}  respectively.

\begin{table}[h]
\centering
\begin{tabular}{ll|ccccc}
Model & Prompt & \multicolumn{4}{c}{Aspect Prediction} & \multicolumn{1}{c}{Title Prediction} \\ 
 & & Idea & Method & Outcome & Future & Title \\ \hline
\multirow{4}{*}{GPT-3.5} & 0-Shot & 0.869 & 0.859 & 0.873 & 0.881 & 0.896 \\ 
 & 2-Shot & 0.874 & 0.870 & 0.875 & 0.879 & \textbf{0.913} \\ 
 & CoT & 0.835 & 0.850 & 0.857 & 0.864 & 0.893 \\ 
 & Few-Shot CoT & 0.866 & 0.856 & 0.862 & 0.872 & 0.904 \\ 
\hline
\multirow{4}{*}{GPT-4} & 0-Shot & 0.871 & \textbf{0.872} & 0.875 & 0.880 & 0.892 \\ 
 & 2-Shot & 0.872 & 0.870 & 0.875 & 0.874 & 0.910 \\ 
 & CoT & 0.869 & 0.865 & \textbf{0.877} & 0.878 & 0.893 \\ 
 & Few-Shot CoT & 0.869 & 0.865 & 0.874 & 0.869 & 0.902 \\ 
\hline
\multirow{4}{*}{Mistral 8x7B} & 0-Shot & 0.869 & 0.869 & 0.875 & 0.881 & 0.884 \\ 
 & 2-Shot & \textbf{0.876} & 0.868 & 0.875 & \textbf{0.882} & 0.897 \\ 
 & CoT & 0.857 & 0.866 & 0.858 & 0.870 & 0.884 \\ 
 & Few-Shot CoT & 0.872 & 0.858 & 0.869 & 0.875 & 0.902 \\ 
\end{tabular}
\caption{Benchmark Results Measured by Cosine Similarity.}
\label{tab:experiment_results_cs}
\end{table}

\begin{table}[h]
\centering
\begin{tabular}{ll|ccccc}
Model & Prompt & \multicolumn{4}{c}{Aspect Prediction} & \multicolumn{1}{c}{Title Prediction} \\ 
 & & Idea & Method & Outcome & Future & Title \\ \hline
\multirow{4}{*}{GPT-3.5} & 0-Shot & 0.839 & 0.845 & 0.855 & 0.860 & 0.875 \\ 
 & 2-Shot & \textbf{0.872} & 0.869 & 0.880 & 0.875 & \textbf{0.892} \\ 
 & CoT & 0.860 & 0.858 & 0.852 & 0.875 & 0.870 \\ 
 & 2-Shot CoT & 0.867 & \textbf{0.875} & 0.878 & \textbf{0.880} & 0.891 \\ 
\hline
\multirow{4}{*}{GPT-4} & 0-Shot & 0.815 & 0.783 & 0.812 & 0.814 & 0.869 \\ 
 & 2-Shot & 0.869 & 0.810 & \textbf{0.886} & 0.854 & 0.883 \\ 
 & CoT & 0.829 & 0.806 & 0.841 & 0.837 & 0.869 \\ 
 & 2-Shot CoT & 0.868 & 0.858 & 0.880 & 0.863 & 0.884 \\ 
\hline
\multirow{4}{*}{Mistral 8x7B} & 0-Shot & 0.823 & 0.822 & 0.840 & 0.838 & 0.822 \\ 
 & 2-Shot & 0.862 & 0.855 & 0.860 & 0.865 & 0.847 \\ 
 & CoT & 0.829 & 0.821 & 0.839 & 0.850 & 0.828 \\ 
 & 2-Shot CoT & 0.870 & 0.866 & 0.875 & 0.877 & 0.862 \\ 
\end{tabular}
\caption{Benchmark Results Measured by BERTScore.}
\label{tab:experiment_results_bs}
\end{table}

\begin{table}[h]
\centering
\begin{tabular}{ll|ccccc}
Model & Prompt & \multicolumn{4}{c}{Aspect Prediction} & \multicolumn{1}{c}{Title Prediction} \\ 
 & & Idea & Method & Outcome & Future & Title \\ \hline
\multirow{4}{*}{GPT-3.5} & 0-Shot & 0.014 & 0.017 & 0.032 & 0.027 & 0.068 \\ 
 & 2-Shot & 0.034 & \textbf{0.029} & 0.042 & 0.033 & \textbf{0.101} \\ 
 & CoT & 0.015 & 0.018 & 0.020 & 0.023 & 0.050 \\ 
 & 2-Shot CoT & 0.026 & 0.025 & 0.031 & 0.027 & 0.079 \\ 
\hline
\multirow{4}{*}{GPT-4} & 0-Shot & 0.008 & 0.006 & 0.012 & 0.009 & 0.049 \\ 
 & 2-Shot & 0.028 & 0.008 & 0.050 & 0.017 & 0.081 \\ 
 & CoT & 0.010 & 0.007 & 0.021 & 0.013 & 0.052 \\ 
 & 2-Shot CoT & 0.025 & 0.019 & 0.041 & 0.016 & 0.064 \\ 
\hline
\multirow{4}{*}{Mistral 8x7B} & 0-Shot & 0.014 & 0.014 & 0.027 & 0.020 & 0.020 \\ 
 & 2-Shot & 0.036 & 0.023 & 0.044 & 0.033 & 0.048 \\ 
 & CoT & 0.014 & 0.014 & 0.023 & 0.020 & 0.011 \\ 
 & 2-Shot CoT & \textbf{0.039} & 0.026 & \textbf{0.056} & \textbf{0.035} & 0.060 \\ 
\end{tabular}
\caption{Benchmark Results Measured by BLEU.}
\label{tab:experiment_results_bleu}
\end{table}

\subsection{URL for Dataset}
\label{app:url}
\begin{itemize}
    \item GitHub Repo (for reproducible results): \url{https://github.com/xingjian-zhang/massw}
    \item Download link for dataset: The download script is provided in the GitHub Repo above. Readers can also download through these links manually. 
        \begin{itemize}
            \item \textbf{MASSW}:\\ \url{https://www.dropbox.com/scl/fi/ykkrpf269fikuchy429l7/massw_v1.tsv?rlkey=mssrbgz3k8adij1moxqtj34ie&dl=1}
            \item \textbf{MASSW Metadata}:\\ \url{https://www.dropbox.com/scl/fi/r2jlil9lj0ypo2fpl3fxa/massw_metadata_v1.jsonl?rlkey=ohnriak63x4ekyli25naajp0q&dl=1}
        \end{itemize}
\end{itemize}

\subsection{Dataset Documentation and Indended Uses}
We use the Data Cards recommended by the submission guideline.\\
Please see \url{https://xingjian-zhang.github.io/massw/}.
\label{sec:doc}

\subsection{Author Statement}
All the authors bear all responsibility in case of violation of rights, etc., and we confirm the data license.

\end{document}